%% file: main.tex
\pdfoutput=1
\documentclass[11pt, letterpaper, shortlabels]{bytedance_seed}

\usepackage{hyperref}
\usepackage{color-edits}
\usepackage{minitoc}
\usepackage[toc,page,header]{appendix}

\newcommand{\eg}{{e.g.}}
\newcommand{\ie}{{i.e.}} 

\ifx\assumption\undefined
\newtheorem{assumption}{Assumption}
\fi

\makeatletter
\def\adl@drawiv#1#2#3{%
        \hskip.5\tabcolsep
        \xleaders#3{#2.5\@tempdimb #1{1}#2.5\@tempdimb}%
                #2\z@ plus1fil minus1fil\relax
        \hskip.5\tabcolsep}
\newcommand{\cdashlinelr}[1]{%
  \noalign{\vskip\aboverulesep
           \global\let\@dashdrawstore\adl@draw
           \global\let\adl@draw\adl@drawiv}
  \cdashline{#1}
  \noalign{\global\let\adl@draw\@dashdrawstore
           \vskip\belowrulesep}}
\makeatother
\newcommand{\method}{Agent-R\xspace}

\usepackage{wrapfig}
\title{\method: Training Language Model Agents to Reflect via Iterative Self-Training}

\usepackage[all]{hypcap}

\usepackage{hyperref}[citecolor=magenta,linkcolor=magenta]

\hypersetup{
    colorlinks = true,
    citecolor = {magenta},
}

\usepackage{multirow, makecell, caption}\usepackage{microtype}
\usepackage{graphicx}
\usepackage{booktabs} 

\usepackage{amsmath}
\usepackage{amssymb}
\usepackage{mathtools}
\usepackage{amsthm}
\usepackage{mathrsfs}
\usepackage{nicefrac}
\usepackage{dsfont}
\usepackage{enumitem}
\usepackage{ulem}
\usepackage{arydshln}
\usepackage{xspace}
\usepackage{subcaption}
\usepackage{wrapfig}
\usepackage{lipsum}
\usepackage{listings}

\usepackage{amsmath}
\usepackage{amssymb}
\usepackage{mathtools}
\usepackage{amsthm}
\usepackage{bbm}

\usepackage{algpseudocode}
\usepackage{setspace}

\usepackage{color}
\definecolor{deepblue}{rgb}{0,0,0.5}
\definecolor{deepred}{rgb}{0.6,0,0}
\definecolor{deepgreen}{rgb}{0,0.5,0}

\newcommand\pythonstyle{\lstset{
basicstyle=\ttfamily\footnotesize,
language=Python,
morekeywords={self, clip, exp, mse_loss, uniform_sample, concatenate, logsumexp},              
keywordstyle=\color{deepblue},
emph={MyClass,__init__},          
emphstyle=\color{deepred},    
stringstyle=\color{deepgreen},
frame=single,                         
showstringspaces=false
}}

\lstnewenvironment{python}[1][]
{
\pythonstyle
\lstset{#1}
}
{}

\newcommand\pythoninline[1]{{\pythonstyle\lstinline!#1!}}

\makeatletter
\def\mathcolor#1#{\@mathcolor{#1}}
\def\@mathcolor#1#2#3{%
  \protect\leavevmode
  \begingroup
    \color#1{#2}#3%
  \endgroup
}
\makeatother

\Crefformat{equation}{#2Eq.\;(#1)#3}

\Crefformat{figure}{#2Figure #1#3}
\Crefformat{assumption}{#2Assumption #1#3}
\Crefname{assumption}{Assumption}{Assumptions}

\usepackage{crossreftools}
\usepackage{dsfont}
\usepackage{nicefrac}

\author[1,2,*,\dagger]{Siyu Yuan}
\author[2,*,\dagger]{Zehui Chen}
\author[1,\dagger]{Zhiheng Xi}
\author[1,2,\dagger]{Junjie Ye} 
\author[2]{Zhengyin Du}
\author[2]{Jiecao Chen}

\affiliation[1]{Fudan University}
\affiliation[2]{ByteDance Seed}

\contribution[*]{Equal contributions}
\contribution[\dagger]{Work done at ByteDance Seed}

\abstract{
Large Language Models (LLMs) agents are increasingly pivotal for addressing complex tasks in interactive and agentic environments. 
Existing work primarily focuses on enhancing performance through behavior cloning from stronger experts, yet such approaches often falter in real-world applications, mainly due to the inability to recover from errors. 
However, step-level critique data is notoriously difficult and expensive to collect. 
Automating and dynamically constructing self-critique datasets is thus crucial to empowering models with intelligent agent capabilities.
In this work, we propose an iterative self-training framework, \textbf{\method}, that enables language \textbf{Agent} to \textbf{R}eflect on the fly. 
Unlike traditional methods that reward or penalize actions solely based on correctness, our approach leverages Monte Carlo Tree Search (MCTS) to construct training samples that recover correct trajectories from erroneous ones. 
A key challenge of agent task reflection lies in the necessity for timely revision rather than waiting until the end of a rollout to revise errors. 
To address this, we introduce a model-guided critique construction mechanism: the actor model identifies the first error step (within its current capability) in a failed trajectory. Starting from it, we splice it with the adjacent correct path, which shares the same parent node in the tree. This strategy enables the model to learn reflection based on its current policy, therefore yielding better learning efficiency.
To further explore the scalability of this self-improvement paradigm, we investigate iterative refinement of both error correction capabilities and dataset construction. 
Our findings demonstrate that this approach continuously improves the model’s ability to recover from errors and enables earlier/timely error correction. 
Extensive experiments on three representative interactive and agentic environments show that the proposed framework effectively equips agents to identify and correct erroneous actions while avoiding loops, achieving superior performance compared to baseline methods (+5.59\%).
}

\date{\today}
\correspondence{Siyu Yuan at \email{syyuan21@m.fudan.edu.cn}}
\checkdata[Project Page]{\url{https://github.com/bytedance/Agent-R}}

\begin{document}

\maketitle
\section{Introduction}\label{sec:intro}
\input{section/intro.tex}
\section{Preliminary}\label{sec:preliminary}
\input{section/preliminary.tex}
\section{Method}\label{sec:method}
\input{section/method.tex}
\section{Experiment}\label{sec:exevalu}
\input{section/exp.tex}

\section{Related Work}\label{sec:related}
\input{section/related.tex}
\section{Conclusion}\label{sec:conclusion}
\input{section/conclusion.tex}

\clearpage

\bibliographystyle{plainnat}
\bibliography{arxiv_paper}
\clearpage

\beginappendix
\input{section/appendix.tex}

\end{document}

%% file: section/intro.tex
Large Language Models (LLMs) have become foundational tools in solving complex tasks across interactive and agentic environments~\citep{wang-etal-2022-scienceworld,yao2022webshop,deng2023mindweb,zhang-etal-2024-timearena,chen2024mindsearch}. 
These LLM-based agents are increasingly employed in scenarios requiring capabilities such as autonomous decision-making, error correction, and task optimization~\citep{prasad-etal-2024-adapt,liu2024agentbench}.
Despite the impressive performance of the existing methods, their reliance on behavior cloning from stronger experts poses significant limitations: 
due to the adoption of all-correct trajectories for training, it struggles to proactively self-correct errors, leading to cascading failures and suboptimal task performance~\citep{zhang2024understanding,xie2024revealing}. 
This limitation arises from an inability to effectively detect errors or revise trajectories dynamically once errors occur, highlighting the need for methods emphasizing timely revision capabilities.

Previous work has proposed methods relying on explicit error signals or reward functions for self-correction. 
However, these methods mainly focus on single-turn scenarios, such as code repair~\citep{kim2024language,chen2024teaching}, tool use~\citep{olausson2023self}, and mathematical reasoning~\citep{kumar2024training,havrilla2024glore}. 
In contrast, tasks in interactive and agentic environments usually involve multi-turn interactions and do not reveal explicit error signals until reaching the terminal state. 
Additionally, unlike mathematical reasoning~\citep{xi2024enhancing}, designing high-quality reward functions to critique intermediate actions in long interactive trajectories remains difficult.
\citet{wang-etal-2024-e2cl} improves agents' self-reflection in embodied tasks, but it heavily relies on expert models for correction during training data construction.

A critical bottleneck in enhancing error recovery in interactive and agentic environments is the lack of step-level reflection data.
Traditional approaches to collecting these datasets involve labor-intensive annotation processes, which are both time-consuming and costly~\citep{lin2024criticbench,zeng2024mr,zheng2024processbench}. 
Without robust reflection data, models face challenges in identifying and correcting their own errors, limiting their utility as intelligent agents. 
Constructing reflection datasets is thus essential for building agents capable of self-reflection and better decision-making.
However, how to automatically construct such training samples is non-trivial. 
A significant challenge of agent task reflection lies in the necessity for timely revision rather than waiting until the end of a rollout to revise errors. 
If corrections are applied only at the end of the trajectory, the delayed revisions prevent agents from learning to detect and address errors as they occur, undermining their capacity for real-time self-reflection. 
Furthermore, delayed revisions may leave catastrophic errors unaddressed, particularly those occurring early in the trajectory.

\begin{figure*}
    \centering
    \includegraphics[width=1.0\linewidth]{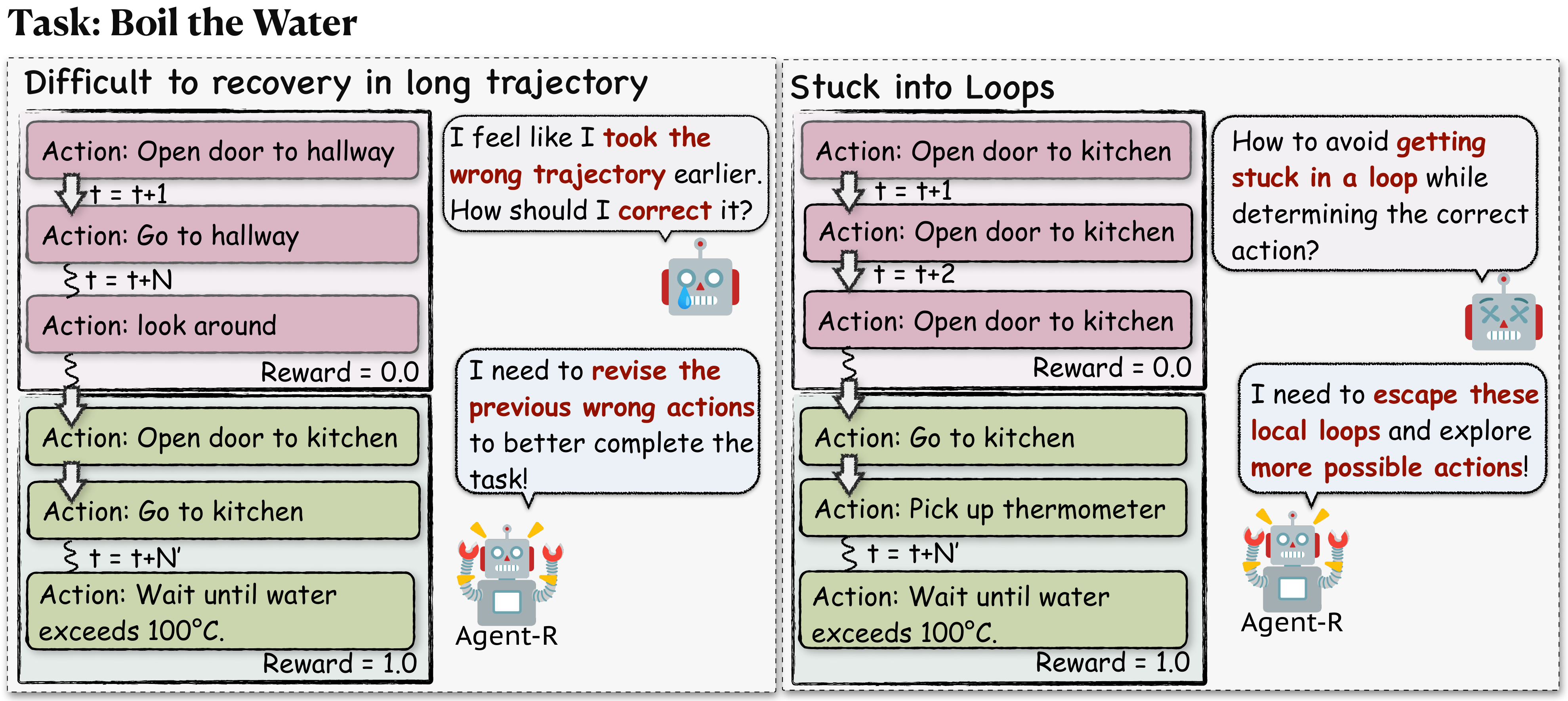}
    \caption{Illustration of language agents struggling with error correction in trajectory generation. These errors can cause agents to enter loops, hindering recovery in long trajectories and resulting in suboptimal outcomes.
    \method enables agents to effectively detect and address errors in real-time, handling long-horizon tasks and avoiding loops with greater self-reflection capabilities.
}
    \label{figue:front}
\end{figure*}

To address these challenges, we propose \textbf{\method}, a novel framework designed to enable LLM-based agents to perform on-the-fly reflection and self-improvement. Unlike previous reward-based approaches, which directly penalize or reward actions based solely on outcome correctness~\citep{song-etal-2024-trial,xiong-etal-2024-watch,shi-etal-2024-direct,putta2024agent}, \method introduces a dynamic self-training framework that revises errors at the step level. 
By leveraging Monte Carlo Tree Search (MCTS)~\citep{kocsis2006bandit}, \method dynamically constructs training samples that recover correct trajectories from erroneous ones, effectively guiding the agent to navigate complex decision spaces.
Specifically, \method identifies the most suitable revision step (based on the current actor model) in an incorrect trajectory and connects it with the subsequent correct trajectory, enabling real-time recovery instead of rolling out to the end of the trajectory.
This dynamic revision process not only enhances the agent’s reflection ability but also mitigates the risk of concatenating inconsistent or incoherent trajectories, which can occur with naive correction strategies.
To further explore the scalability of \method, we iteratively refine both its error correction capabilities and its dataset construction process. 
This iterative approach enables the agent to continuously improve its performance, correcting errors earlier in trajectories and reducing the likelihood of entering erroneous loops.

Extensive experiments across three diverse and representative interactive environments~\citep{wang-etal-2022-scienceworld,yao2022webshop,prasad-etal-2024-adapt} demonstrate that \method significantly outperforms baselines by providing language agents with self-correction capability similar to human decision-making.
Notably, training with revision trajectories surpasses the use of expert trajectories.
Additionally, \method enables agents to more effectively identify and correct erroneous actions while avoiding loops, a limitation observed in previous methods trained on expert trajectories.
Overall, the contributions of this paper can be summarized as follows:
\begin{itemize}
	\item We are the first to highlight the critical challenge of error correction in interactive environments for language agents.
    This limitation is overlooked by prior methods.
	\item We introduce \method, a novel iterative framework that empowers language agents with self-reflection capabilities in interactive environments. By leveraging MCTS and a model-guided reflection mechanism, \method dynamically constructs and revises trajectories, enabling agents to improve their behavior through autonomous self-correction without relying on human or expert model supervision.
	\item We conduct extensive experiments across three diverse interactive environments, demonstrating that \method significantly outperforms baselines. By leveraging self-correction, \method enables agents to effectively detect and address errors in real-time, avoiding loops and handling long-horizon tasks with greater robustness.
\end{itemize}

%% file: section/preliminary.tex
\subsection{Task Formulation}
In this paper, we focus on tasks with partial observations in interactive environments. 
Following prior work~\citep{song-etal-2024-trial,qiao2024agent}, these tasks can be formulated as a Partially Observable Markov Decision Process (POMDP): $(\mathcal{U}, \mathcal{S}, \mathcal{A}, \mathcal{O}, \mathcal{T}, \mathcal{R})$.
The instruction space $\mathcal{U}$ provides task descriptions and their corresponding requirements. $\mathcal{S}$ represents the state space, $\mathcal{A}$ is the action space, and $\mathcal{O}$ is the observation space. The transition function $\mathcal{T}: \mathcal{S} \times \mathcal{A} \rightarrow \mathcal{S}$ is determined by the environment, while the reward function $\mathcal{R}: \mathcal{S} \times \mathcal{A} \rightarrow [0, 1]$ specifies the reward.
For language agent, $\mathcal{U}$, $\mathcal{S}$, $\mathcal{A}$, and $\mathcal{O}$ are represented in natural language.
At each time step $t$, the historical trajectory $\tau_t$ is defined as: 
\begin{align} 
\tau_t = (a_1, o_1, \dots, a_t, o_t) \sim \pi_{\theta}(\tau_t|u)
\end{align} 
where $(a_1, o_1, \dots, a_t, o_t)$ is a sequence of actions and observations, and $o_t$ is the observation obtained after executing action $a_t$.

The language agent with parameter $\theta$ is the actor model $\pi_\theta$. 
At each time step $t+1$, the actor model should generate $a_{t+1}$ based on $\tau_t$ and instruction $u$, \ie, $a_{t+1} \sim \pi_\theta(\cdot | \tau_t, u)$. 
We adopt the ReAct approach~\citep{yao2023react}, generating rationales before each action to enhance performance.
The task ends when it is successfully completed, or the maximum number of rounds is reached. 
Then, the final reward  $r(\tau) \in [0, 1]$ is then given by the interactive environments. 
Thus, at the terminal time step $T$, the probability distribution of the entire trajectory $\tau_T$ is represented as: 
\begin{align}
    \pi_\theta(\tau_T | u) = \prod_{t=1}^T \pi_\theta(a_{t} | \tau_{t-1},u).
\end{align}

\subsection{Monte Carlo Tree Search}
Monte Carlo Tree Search (MCTS) is a decision-making algorithm widely used in complex decision processes~\citep{kocsis2006bandit,browne2012survey}. It builds a search tree and simulates outcomes to estimate the value of actions.  
MCTS operates by iteratively developing a decision tree through four key stages:  
\begin{itemize}  
    \item \textbf{Selection}: Uses the Upper Confidence bound for Trees (UCT) strategy~\citep{kocsis2006bandit} to select the next node for expansion.  
    \item \textbf{Expansion}: Based on the selected node, generates and adds new nodes to the tree. 
    \item \textbf{Simulation}: Performs multiple rollouts from the newly added node to a terminal node, which could be either a terminal state or a predetermined maximum tree depth.  
    \item \textbf{Backpropagation}: Updates node values based on the results of the simulations.
\end{itemize}

%% file: section/method.tex
\begin{figure*}[t]
    \centering
    \includegraphics[width=\linewidth]{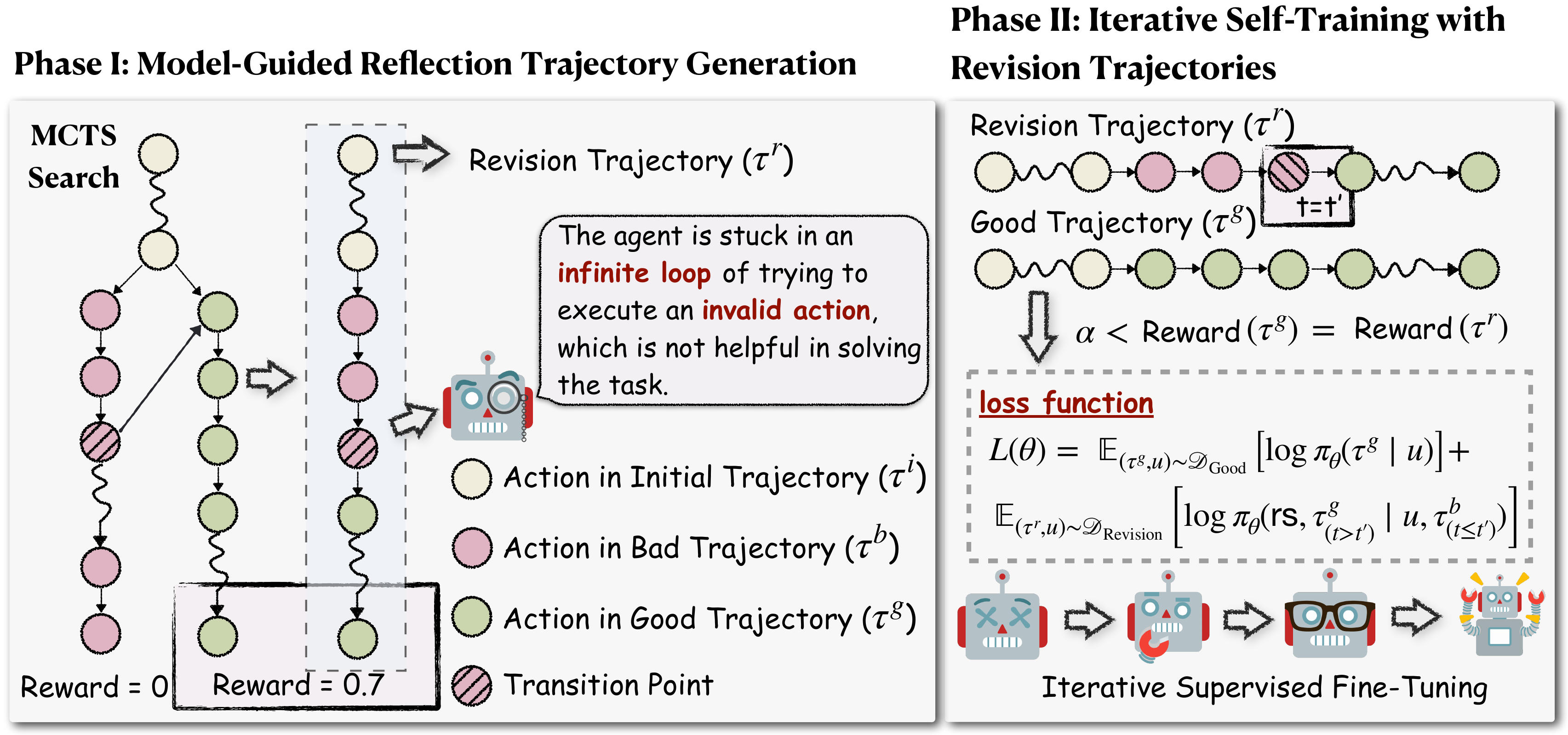}
    \caption{The framework of \method consists of two phases. In Phase I, we adopt MCTS and a model-guided reflection mechanism to construct revision trajectories. In Phase II, the agents are trained using the collected revision trajectories. These two phases can be repeated iteratively. $\texttt{rs}$ is the revision signal, $t'$ is the transition point between the bad and good trajectories, and $L(\theta)$ is the loss function to be optimized.}
    \label{fig:method}
\end{figure*}

In this section, we present \method in detail, which consists of two core phases, as illustrated in Figure~\ref{fig:method}.
Phase I: Model-Guided Reflection Trajectory Generation ($\mathsection$~\ref{sec:phaseI}), which utilizes Monte Carlo Tree Search (MCTS) to dynamically generate revision trajectories, transforming erroneous trajectories into corrected ones by identifying the most suitable reflection step (for current policy model). 
Phase II: Iterative Self-Training with Revision Trajectories, where agents are iteratively trained on the dynamically constructed revision trajectories. 
Such an iterative training manner enables us to explore the scalability of \method, where agents progressively learn harder (earlier) revision steps based on its current policy and improve their decision-making process, avoiding cascading errors and loops.

\subsection{Phase I: Model-Guided Reflection Trajectory Generation}\label{sec:phaseI}

\paragraph{Reflection Trajectory Definition}
We first define four types of trajectories that play a central role in \method: initial trajectories, bad trajectories, good trajectories, and revision trajectories. These are described as follows\footnote{The detailed formal definitions and mathematical formulations for each type of trajectory are shown in Appendix~\ref{appendix:definition}}:
\begin{itemize}[leftmargin=*]
    \item \textbf{Initial Trajectory}, denoted as $\tau^{i}= (a_{1}^{i}, o_{1}^{i}, \dots, a_{t}^{i}, o_{t}^{i})$:
    An initial trajectory represents the initial sequence of actions and observations.
    \item \textbf{Bad Trajectory}, denoted as $\tau^{b}= (\tau^{i}, a_{t+1}^{b}, o_{t+1}^{b}, \dots, a_{T_{b}}^{b}, o_{T_{b}}^{b})$:  
    A bad trajectory extends the initial trajectory $\tau^{i}$ with a sequence of suboptimal actions and observations, leading to an erroneous or less-rewarding outcome.
    \item \textbf{Good Trajectory}, denoted as $\tau^{g}= (\tau^{i}, a_{t+1}^{g}, o_{t+1}^{g}, \dots, a_{T_{g}}^{g}, o_{T_{g}}^{g})$:  
    A good trajectory builds upon the initial trajectory $\tau^{i}$, involving optimal or high-reward actions and observations.
    \item \textbf{Revision Trajectory}, denoted as $\tau^{r}$: 
    A revision trajectory is constructed by correcting a bad trajectory with a good trajectory. Formally, it is defined as:
    \begin{equation}
    \tau^{r} = (\tau^{i}, a_{t+1}^{b}, o_{t+1}^{b}, \dots, a_{t'}^{b}, o_{t'}^{b}, \texttt{rs}, a_{t+1}^{g}, o_{t+1}^{g}, \dots, a_{T_{g}}^{g}, o_{T_{g}}^{g}),
    \end{equation}
    where $t'$ represents the transition point between the bad and good trajectory segments, and $\texttt{rs}$ is the revision signal marking the transition. The revision signal $\texttt{rs}$ is a brief communication such as: \texttt{Assistant: [reflection on the error] \textbackslash n Human: OK.}, which facilitates the agent's reflection on its past actions.

\end{itemize}

The interactive environment will provide the final reward after the trajectory has terminated.
Based on the final reward, we define constraints to ensure that bad trajectories can be effectively corrected into good ones and that the quality of the revision trajectory matches that of the good trajectory. The reward conditions are as follows:
\begin{equation}
\label{eq:reward-constraints}
r(\tau^{b}) < \beta < r(\tau^{g}) \leq 1, \quad \alpha < r(\tau^{g}) = r(\tau^{r}),
\end{equation}
where $\beta$ is a threshold separating the rewards of bad and good trajectories, and $\alpha$ represents a lower bound for high-quality trajectories, encouraging revisions that are of consistently high quality.
$r(\tau^{g}) = 1$ means that this good trajectory is a \textbf{Optimal Trajectory}.

\paragraph{Trajectory Collection with MCTS}

To efficiently explore and reflect errors within the trajectory space, we employ Monte Carlo Tree Search (MCTS) to collect revision trajectories. 
This method systematically searches through possible action paths, ensuring diversity in the generated trajectories while balancing exploration and exploitation.

Starting from the initial root node $s_0$, which corresponds to the user-provided instruction $u$, MCTS iteratively performs four key stages: selection, expansion, simulation, and backpropagation.
As shown in Figure~\ref{fig:method}, during the simulation stage, a default rollout policy is used to sample future actions. 
To improve reward estimation and ensure diversity in collected trajectories, multiple rollouts are performed for each simulation.
In the selection phase, we balance exploration and exploitation using the UCT criterion:
\begin{equation}
\label{eq:uct}
UCT(s) = Q(s) + c_{\texttt{uct}} \cdot \sqrt{\frac{\log N_p(s)}{N(s)}},
\end{equation}
where $Q(s)$ is the average reward of state $s$, $N(s)$ the number of visits to state $s$, $N_p(s)$ is the total visit count of the parent node of $s$, and $c_{\texttt{uct}}$ is a constant that controls the exploration-exploitation trade-off.

When a terminal node is reached, either due to a terminal state or exceeding a predefined maximum depth, MCTS generates a trajectory from the root node to the terminal node and obtains the final reward for the trajectory from the environment.
By performing multiple rollouts, we collect a diverse set of trajectories.  
Good and bad trajectories share the same initial trajectory and begin to diverge after step $t$, where their quality is determined by the final reward given by environments.
Good trajectories achieve higher rewards, while bad trajectories fall below a threshold.  

To construct the revision trajectories, we first manually design ten different revision thoughts for \texttt{rs}.\footnote{The ten revision thoughts are listed in Appendix~\ref{appendix:prompt}.} 
During the construction process, one of these thoughts is randomly sampled to connect the segments of bad and good trajectories. 
Following \cite{kumar2024training}, a simple baseline, referred to as the \textbf{Direct-Revision Trajectory}, sets the transition point $t'$ as the terminal step of the bad trajectory ($t' = T_b$), meaning that the good trajectory is directly appended to the end of the bad trajectory.

While this approach is simple, it omits a key distinction of agent task reflection: timely revision is necessary for most agent tasks instead of waiting until the end of a rollout to revise errors.
If corrections are applied only at the end of the trajectory, the agent cannot self-reflect and learn to detect and reflect errors as they occur. 
Moreover, the direct-revision trajectory fails to address catastrophic errors that may occur early in the trajectory. 

\paragraph{Transition Point Determination with Actor Model}
To address the above problem, we propose a model-guided revision trajectory construction approach inspired by human cognitive processes. 
In this approach, the language agent evaluates each action within its self-generated bad trajectories to identify errors based on the actor model's current capability, which is in line with the reflection on its actions. 
Specifically, when the agent detects an incorrect action $a_{t^*}$, the transition point is set to $t' = t^*$, and the bad trajectory is truncated at this point. 
The agent then constructs a revision trajectory by combining the corrected portion of the bad trajectory with the corresponding segment of the good trajectory.

This approach is fundamentally interactive and leverages the agent's self-awareness to actively refine its behavior in real-time. 
Figure~\ref{fig:ablation} in our experiments demonstrates the superiority of such transition point determination over the direct-revision approach. 
The prompt used for the model-guided transition point determination is provided in Appendix~\ref{appendix:prompt}.
It not only ensures that the revisions are grounded in the agent's own learned dynamics but also lays the foundation for scalable self-improvement from weaker to stronger behaviors (discusses in \S \ref{sec:phaseII}).

\subsection{Phase II: Iterative Self-Training with Revision Trajectories}\label{sec:phaseII}

In this phase, the goal is to train language agents using self-generated revision trajectories collected through MCTS. 
The agent uses its own rollouts to identify and correct errors dynamically through self-reflection, progressively improving its policy based on its own experiences during these rollouts.

Although training on revision trajectories enables the agent to develop self-reflection capabilities, relying solely on these trajectories may initially hinder the agent's ability to identify the correct trajectory.
A common solution in previous works is to train the model using both optimal and revision trajectories.
However, this still suffers from the cold-start problem, where the agent begins with little to no knowledge of the environment and must depend on trial and error. 
This results in a limited number of optimal trajectories discovered early in training.

To address this issue, we propose mixing revision trajectories with good trajectories during training. 
Over the course of self-training, we gradually increase the value of $\alpha$ in Equation~\ref{eq:reward-constraints}, which encourages the good trajectories to progressively converge toward optimal trajectories. 
This enables the agent to improve both its error reflection capabilities and its ability to identify the correct trajectory over time.

Additionally, following the strategy from AgentTuning \citep{zeng-etal-2024-agenttuning}, we combine agent trajectory datasets with general datasets, such as ShareGPT\footnote{\url{https://huggingface.co/datasets/anon8231489123/ShareGPT_Vicuna_unfiltered}}, to improve generalization capabilities. 
This approach has been shown to outperform training exclusively on agent tasks and enhances the agent's ability to generalize across a broader range of environments. 
We adopt this strategy in our framework.
Specifically, for the base model $\pi_\theta$, the objective is to find the optimal policy $\pi_\theta(\tau\mid u)$ that minimizes the loss function $L(\theta)$:
\begin{equation}
\label{eq:training-loss}
\begin{aligned}
L(\theta)=&\; \eta\cdot \{\mathbb{E}_{(\tau^g, u) \sim \mathcal{D_{\mathrm{Good}}}}
\left[\log \pi_\theta(\tau^g \mid u)\right] + \mathbb{E}_{(\tau^r, u) \sim \mathcal{D_{\mathrm{Revision}}}}
\left[\log \pi_\theta(\texttt{rs}, \tau^g_{(t>t')} \mid u, \tau^b_{(t \le t')})\right]\} \\
&+ (1-\eta)\cdot\mathbb{E}_{(x, y) \sim \mathcal{D_{\mathrm{general}}}}
\left[\log \pi_\theta(y \mid x)\right],
\end{aligned}
\end{equation}
where $(x,y)$ is the input-output pair from the general dataset, $\eta$ is the mixture ratio between $\mathcal{D_{\mathrm{Revision}} + D_{\mathrm{Good}}}$ and $\mathcal{D_{\mathrm{general}}}$ and $\log \pi_\theta(\tau \mid u) = \sum_{t=1}^{T}\log \pi_\theta(a_t \mid \tau_{t-1}, u)$. 

To further explore the stability of our approach, we iteratively refine both its error correction capabilities and its dataset construction process. Specifically, at each iteration, we re-collect model-guided revision trajectories based on the current actor model and perform SFT to optimize Equation~\ref{eq:training-loss}. 
As the agent trains over multiple iterations, it builds enhanced self-reflection capabilities and progressively transforms from weaker, error-prone behaviors to stronger, more efficient performance in interactive environments.
This process supports the agent's ability to dynamically recover from errors and learn from its experiences, facilitating its reflection on actions and continuous performance enhancement.

%% file: section/exp.tex
In this section, we conduct extensive experiments on three interactive environments to demonstrate the effectiveness of \method.
\input{table/statistics}

\subsection{Interactive and Agentic Environments}
Following previous work~\citep{xi2024agentgym,prasad-etal-2024-adapt}, we conduct experiments on three types of representative interactive environments: 
\begin{itemize}
    \item \textbf{WebShop}~\citep{yao2022webshop}, which is an interactive web environment for online shopping. 
    It contains 12k instructions and offers over one million real products from amazon.com. 
    Agents can click buttons on the webpage or perform searches using the search engine.
    \item \textbf{ScienceWorld}~\citep{wang-etal-2022-scienceworld}, which is a scientific, text-based environment designed to evaluate agents' scientific reasoning abilities. 
    It includes 30 types of scientific tasks at the standard elementary science curriculum level.
    \item \textbf{TextCraft}~\citep{prasad-etal-2024-adapt}, which is a text-based environment for crafting Minecraft items. 
    It constructs a crafting tree based on Minecraft’s recipes. 
    Each task provides a target item and a list of crafting commands generated by the tree. 
    Agents receive a reward of 1 when they successfully craft the target item.
\end{itemize}

\subsection{Experiment Setting}
We conduct our experiments on AgentGym~\citep{xi2024agentgym}, an interactive platform that includes diverse agent environments.

\paragraph{Data Split}
The statistical details of these three interactive environments are shown in Table~\ref{tab:statistics}. 
To create the revision trajectories, we randomly sample 300 simulations from WebShop, 200 from SciWorld, and 200 from TextCraft to conduct MCTS.
We set the distinguishable gap $\beta = 0.2$ between bad and good trajectories in Equation~\ref{eq:reward-constraints}. 
As mentioned in $\mathsection$~\ref{sec:phaseI}, considering the cold-start problem for agents in these challenging environments, the initial threshold for good trajectories is set relatively low and gradually increases in later iterations. 
Specifically, we conduct \method for three iterations. We set $\alpha = 0.5$ for iteration 1, $\alpha = 0.7$ for iteration 2, and $\alpha = 1.0$ for iteration 3. By iteration 3, the good trajectories converge toward optimal trajectories.  
Following AgentTuning~\citep{zeng-etal-2024-agenttuning}, we use ShareGPT as the general dataset $D_{\texttt{general}}$ and set $\eta = 0.2$ in Equation~\ref{eq:training-loss}.
Following \cite{xi2024agentgym}, we select 200 tasks for the WebShop test set, 200 for the SciWorld test set, and 100 for the TextCraft test set.

\paragraph{MCTS Settings}  
For all simulations in the three interactive environments, during the trajectory self-generation stage, we sample $k = 8$ rollouts for each Monte Carlo estimation. The depth $d$ is set to 20 for all tasks.  
In the expansion phase, the temperature of the LLMs is set to 1. At each depth, the LLMs generate 4 candidate actions as new child nodes.  
In the selection phase, to balance exploration and exploitation, we set $c_{\texttt{uct}} = 0.25$ for UCT in Equation~\ref{eq:uct}.

\paragraph{Training Settings}  
We perform \method for three iterations and present the results of the third iteration in Table~\ref{tab:overall}. 
We set the epoch number to 3 for the first iteration and 1 for subsequent iterations to avoid over-fitting.  
Detailed training settings are provided in Appendix~\ref{appendix:parameters}.
Our main backbone model is the instruct version of LLama-3.1-8B, \ie, \texttt{Llama-3.1-8B-Instruction}~\citep{dubey2024llama}.

\paragraph{Baselines}  
Following \cite{xi2024agentgym}, we select closed-source models, \ie, GPT-3.5-Turbo~\citep{openai2022chatgpt}, GPT-4-Turbo~\citep{OpenAI2023GPT4}, GPT-4o (\texttt{gpt-4o-2024-08-06})~\citep{OpenAI2023GPT4}, Claude 3~\citep{Anthropic2024Claude3}, and DeepSeek-Chat~\citep{liu2024deepseek}.  
We also select open-source models like Llama2-Chat~\citep{Hugo2023LLaMa2} and agents trained on expert trajectories, \ie, AgentLM~\citep{zeng-etal-2024-agenttuning} and Agent-Flan~\citep{chen-etal-2024-agent}.
Additionally, we compare ETO~\citep{song-etal-2024-trial}, which first applies SFT to a base agent using behavioral cloning and then uses DPO~\citep{rafailov2023direct} to fine-tune the model with contrastive pairs of good and bad trajectories.
We also report the results of Llama-3.1-8B-Instruct trained on direct-revision trajectories mixed with good trajectories over three iterations for comparison (w/ Direct-Revision).

\paragraph{Evaluation Metrics}
Following \cite{xi2024agentgym}, for SciWorld and WebShop, we use the average final reward as the evaluation metric. 
For TextCraft, we use the success rate as the evaluation metric. 
In all three environments, the maximum number of rounds is set to 100.
Detailed evaluation settings are provided in Appendix~\ref{appendix:parameters}.

\input{table/overall}
\subsection{Main Result}
The overall results for the three interactive environments are shown in Table~\ref{tab:overall}. 
We find that:
\begin{enumerate}[leftmargin=*] 
    \item \method significantly improves the performance of language agents in interactive environments. This approach outperforms both advanced closed-source models (\eg, GPT-4o) and agents trained on expert trajectories (\eg, AgentLM and Agent-Flan). The results highlight the critical role of revising erroneous trajectories during the learning process rather than solely relying on expert data.
    \item Trajectories constructed by \method, through dynamic self-reflection and early error detection, lead to faster recovery and more stable learning. This early intervention prevents the propagation of errors, resulting in higher-quality models and better performance than direct-revision methods.
    \item Although adopting contrastive learning methods (\eg, ETO) improves language agent performance in interactive environments, these methods do not equip agents with self-reflection capabilities. This limitation is analyzed in $\mathsection$~\ref{sec:analysis}.  
\end{enumerate}
Additional case study analyses for \textit{\method} are provided in Appendix~\ref{appendix:case_study}, illustrating specific instances of error recovery and trajectory correction.

\subsection{Findings with Analysis}\label{sec:analysis}
\begin{figure*}[t]
    \centering
\includegraphics[width=0.95\linewidth]{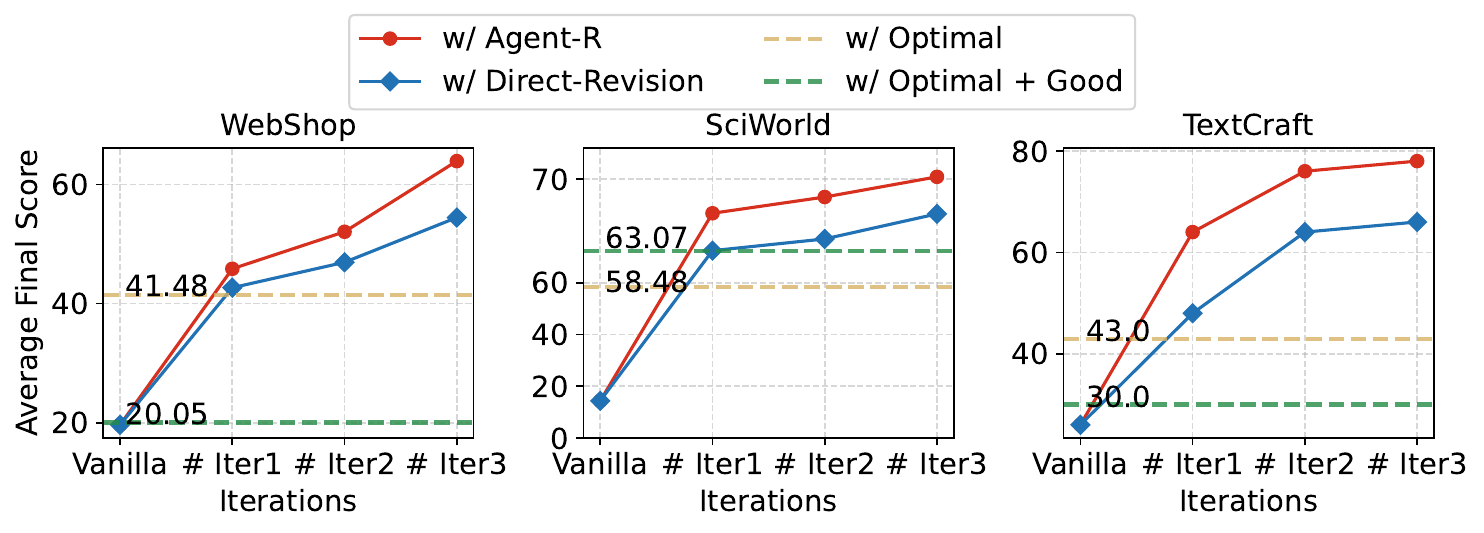}
    \caption{Results of different training trajectories under different iterations on three interactive environments.}
    \label{fig:ablation}
\end{figure*}

\paragraph{\textit{Finding 1: Training with trajectories from \method can outperform using optimal trajectories.}}  

To further investigate the significance of revision trajectories constructed by \method, we compare our approach with Llama-3.1-8B-Instruct trained on direct-revision trajectories combined with good trajectories (\textbf{w/ Direct-Revision Trajectory}) for each iteration. 
Additionally, we include two other ablated variants:
\textbf{w/ Optimal Trajectory}: Llama-3.1-8B-Instruct trained only on self-generated optimal trajectories ($r(\tau)=1$). 
\textbf{w/ Optimal + Good Trajectory}: Llama-3.1-8B-Instruct trained on optimal trajectories combined with the good trajectories from iteration 1 in \method.
The results in Figure~\ref{fig:ablation} show that:  
\begin{enumerate}[leftmargin=*]  
    \item Iterative SFT with trajectories from \method gradually enhances the model's capabilities. It is worth noting that this process relies entirely on self-play to collect and construct trajectories, offering insights and potential future directions for the autonomous evolution of language agents.
    \item Mixing revision trajectories with optimal trajectories in the training process significantly improves performance. This highlights the importance of iterative reflection and error recovery in dynamic environments.
    \item Adding only good trajectories to optimal ones decreases performance. This may be because good trajectories introduce noise, as it is not always completely correct ($r(\tau) \leq 1$ ). Although revision trajectory also contains noise, it provides information on how to recover from errors. This additional information offsets the noise and improves performance.  
\end{enumerate}

\input{table/revision}

\begin{figure*}[t]
    \centering
\includegraphics[width=0.9\linewidth]{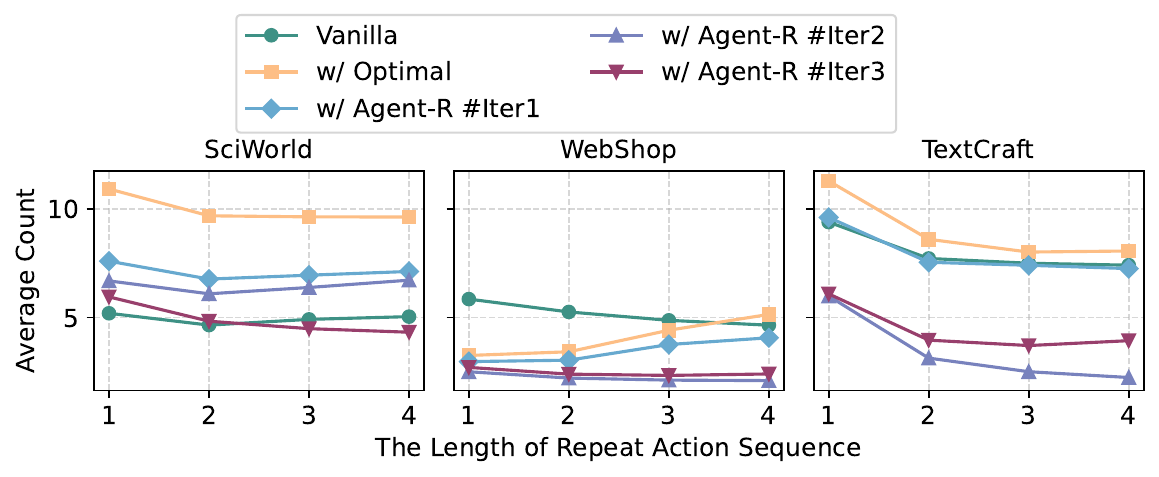}
    \caption{Average count of repeated action lengths for different training trajectories and different iterations in three interactive environments.}
    \label{fig:loop}
\end{figure*}

\paragraph{\textit{Finding 2: \method can effectively provide language agents with self-reflection capabilities.}}
To further explore the self-reflection capabilities of language agents trained with \method, we first collect all failure trajectories ($r(\tau)=0$) from the test set results of Llama-3.1-8B-Instruct.  
Then, given these failure trajectories, we randomly select a time step $t$, truncate the trajectories at that time, and then ask language agents to continue generating actions after $t$ and evaluate whether they can correct these failures to achieve better final rewards.
In this revision setting, the maximum number of rounds for all three environments is set to 50. 
We compare \method with self-generated optimal trajectories ($r(\tau)=1$,  w/ Optimal), GPT-4o-generated optimal trajectories (w/ Optimal+), contrastive pairs of good and bad trajectories (w/ ETO) and direct-revision mixed with good trajectories.  
The results in Table~\ref{tab:revision} indicate that:  
\begin{enumerate}[leftmargin=*] 
    \item Training with \method revision trajectories outperforms direct-revision methods, emphasizing the importance of incorporating early reflection during the trajectory rather than waiting until the end. This on-the-fly correction mechanism is crucial for timely error recovery.
    \item Although training on expert trajectories using SFT or with contrastive pairs of good and bad trajectories using DPO can significantly enhance agent performance in interactive environments, it does not provide satisfactory self-reflection capabilities. If an agent gets stuck in a loop or executes incorrect actions, it cannot identify and correct the error.  
    \item While GPT-4o performs worse than AgentLM and AgentEvol on the test set, it shows better self-reflection capabilities, demonstrating its strong general abilities.
\end{enumerate}

\begin{wrapfigure}{r}{0.5\linewidth}
    \centering
\includegraphics[width=\linewidth]{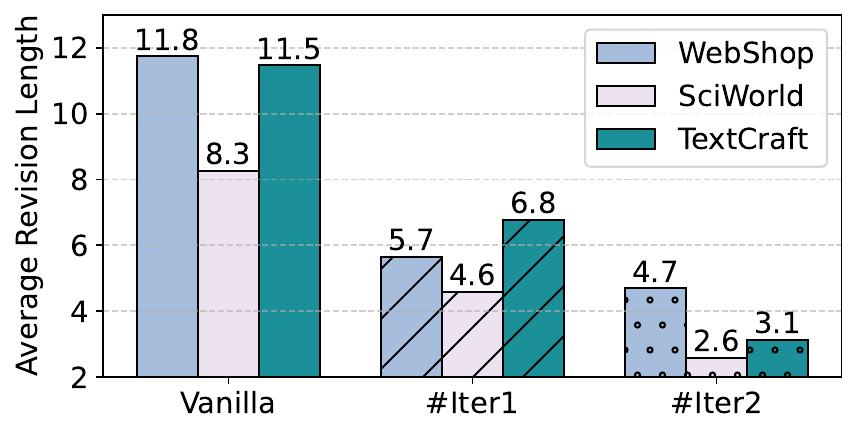}
    \caption{Average revision length of different iterations on three interactive environments.}
    \label{fig:revision}
\end{wrapfigure}

\paragraph{\textit{Finding 3: Training with revision trajectories helps agents more easily identify and correct erroneous actions.}}  
As mentioned in $\mathsection$~\ref{sec:phaseI}, to construct revision trajectories, we ask language agents to identify the first erroneous action in the bad trajectories and then concatenate the correct trajectories after this point.  
To determine if agents trained with iterative SFT can more effectively and quickly detect the first error, we measure the \textbf{revision length}. This is the number of actions from the start of the bad trajectory to the first identified error. We report the average revision length for different iterations across the three interactive environments.  

The results in Figure~\ref{fig:revision} indicate that \method significantly improves the agent's ability to recognize when an error occurs within a trajectory.
Additionally, the shorter average revision length indicates that training with revised trajectories from \method, where error correction is integrated dynamically and earlier in the trajectory, enables the model to more efficiently adjust its path. 
Moreover, this improvement reflects the enhanced reflection ability of the actor model, which could better serve as a critical component to assist other models, a direction worth exploring in future work.

\paragraph{\textit{Finding 4: Training with revision trajectories helps agents avoid getting stuck in loops.}}  
To understand why revision trajectories are more effective than self-generated optimal ones, we measure the \textbf{length of repeat action sequences} in test set trajectories under different training settings. We report the average count, which reflects how often agents repeat the same action sequence when stuck in a loop. For example, a sequence length of 2 with an average count of 5 means the agent repeats the same sequence of two actions 5 times on average.

The results in Figure~\ref{fig:loop} show that although agents can obtain optimal trajectories through MCTS, these trajectories may contain repeated or noisy middle actions, causing the agent to get stuck in dead loops, which negatively impacts performance.  
Compared to training with optimal trajectories, training with trajectories from \method significantly reduces the occurrence of dead loops. This indicates that agents trained with revision trajectories are more likely to explore new actions and avoid getting trapped in local loops. These findings emphasize the importance of revision trajectories in enhancing exploration and overall performance.

\begin{wrapfigure}{r}{0.5\linewidth}
    \centering
\includegraphics[width=\linewidth]{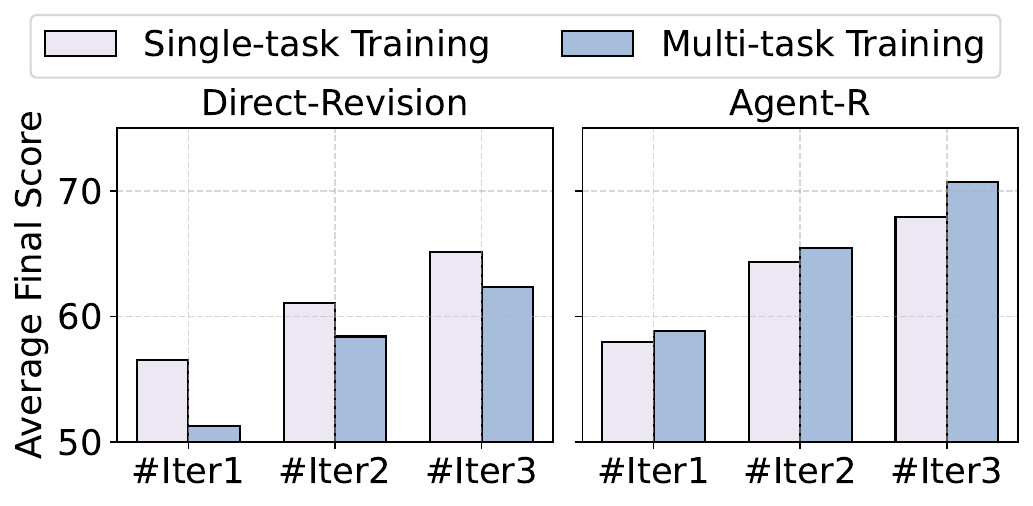}
    \caption{Comparison of different training methods on three interactive environments.}
    \label{fig:multi}
\end{wrapfigure}

\paragraph{\textit{Finding 5: Multi-task training is a more effective strategy for \method.}}  
We compare our multi-task training approach with single-task training, where good and revised trajectories are collected separately for each task using \method.  
The results in Figure~\ref{fig:multi} show that multi-task training is a more effective way to train language agents in interactive environments. Additionally, the trajectories collected by \method are better suited for multi-task training, further enhancing performance.
The detailed results of three environments with further analysis are shown in Appendix~\ref{appendix:mix}.

%% file: table/statistics.tex
\begin{table*}[t]
    \centering
    \small
    \caption{Statistics of three agent interactive environments. Due to different selection thresholds, MCTS can collect different numbers of revision data and good data for the training phase.}
    \label{tab:mcts_results}
    \begin{tabular}{lcccccccc}
        \toprule
        \multirow{2}{*}{\textbf{Dataset}} & \multirow{2}{*}{\textbf{Simulation \#}} & \multicolumn{2}{c}{\textbf{MCTS w/ Iter 1}} & \multicolumn{2}{c}{\textbf{MCTS w/ Iter 2}} & \multicolumn{2}{c}{\textbf{MCTS w/ Iter 3}} & \multirow{2}{*}{\textbf{Test Size}} \\
        \cmidrule(lr){3-4} \cmidrule(lr){5-6} \cmidrule(lr){7-8}
        & & \textbf{Revision} & \textbf{Good} & \textbf{Revision} & \textbf{Good} & \textbf{Revision} & \textbf{Good} & \\
        \midrule
        WebShop    & 300 & 5500 & 600  & 8500 & 1800 & 9000 & 2000 & 200 \\
        SciWorld   & 200 & 8000 & 2500 & 4200 & 1800 & 5400 & 1200 & 200 \\
        TextCraft  & 200 & 6000 & 2000 & 7000 & 3200 & 8000 & 4000 & 100 \\
        \bottomrule
    \end{tabular}
    \label{tab:statistics}
\end{table*}

%% file: table/overall.tex
\begin{table}[t]
    \centering
    \small
    \begin{tabular}{lcccc}
    \toprule
        \textbf{Method} & \textbf{WebShop} & \textbf{SciWorld} & \textbf{TextCraft} & \textbf{Average} \\ 
    \midrule        
    \multicolumn{5}{c}{\textit{Close-sourced Models}} \\
    \midrule
        DeepSeek-Chat~\citep{liu2024deepseek} & 11.00 & 16.80 & 23.00 & 16.93 \\ 
        Claude-3-Haiku~\citep{Anthropic2024Claude3} & 5.50 & 0.83 & 0.00 & 2.11 \\ 
        Claude-3-Sonnet~\citep{Anthropic2024Claude3} & 1.50 & 2.78 & 38.00 & 14.09 \\ 
        GPT-3.5-Turbo~\citep{openai2022chatgpt} & 12.50 & 7.64 & 47.00 & 22.38 \\ 
        GPT-4-Turbo~\citep{OpenAI2023GPT4} & 15.50 & 14.38 & \underline{77.00} & 35.63 \\ 
        GPT-4o~\citep{OpenAI2023GPT4} & 25.48 & 46.91 & 64.00 & 45.46 \\
        \midrule
    \multicolumn{5}{c}{\textit{Open-sourced Models}} \\
    \midrule
        Llama2-Chat-13B~\citep{Hugo2023LLaMa2} & 1.00 & 0.83 & 0.00 & 0.61 \\ 
        AgentLM-7B~\citep{zeng-etal-2024-agenttuning} & 36.50 & 2.75 & 0.00 & 13.08 \\ 
        AgentLM-13B~\citep{zeng-etal-2024-agenttuning} & 39.50 & 10.68 & 4.00 & 18.06 \\ 
        Agent-FLAN~\citep{chen-etal-2024-agent} & 40.35 & 28.64 & 16.00 & 28.33 \\
        \midrule
        Llama-3.1-8B-Instruct~\citep{dubey2024llama} & 19.65 & 14.36 & 26.00 & 20.00 \\ 
        w/ ETO~\citep{song-etal-2024-trial} & 52.80 & \underline{67.55} & 75.00 & \underline{65.12} \\ 
        w/ Direct-Revision & \underline{54.44} & 66.65 & 66.00 & 62.36 \\
        \cdashlinelr{1-5}
        w/ \method & \textbf{63.91} & \textbf{70.23} & \textbf{78.00} & \textbf{70.71} \\ 
    \bottomrule
    \end{tabular}
    \caption{Results of three interactive environments. We train Llama-3.1-8B on revision trajectories collected from \method for three iterations and compare its performance with various models.}
    \label{tab:overall}
\end{table}

%% file: table/revision.tex
\begin{table}[t]
    \centering
    \small
    \begin{tabular}{lcccc}
    \toprule
        \textbf{Method} & \textbf{WebShop} & \textbf{SciWorld} & \textbf{TextCraft} & \textbf{Average} \\ 
    \midrule
        GPT-4o & 24.10 & 35.56 & 8.22 & 22.63   \\ 
    \midrule
        AgentLM-7B &  20.96 & 0.00 & 0.00 & 6.99   \\ 
        AgentLM-13B & 23.20 & 0.22 & 0.00 & 7.81   \\ 
        Agent-FLAN & 19.37 & 0.00 & 0.00 & 6.46  \\ 
        
    \midrule
    \multicolumn{5}{c}{\textit{Llama-3.1-8B}} \\
    \midrule
        w/ Optimal & 31.38 & 1.15 & 0.00 & 10.84   \\ 
        w/ Optimal+ & 25.03 & 0.00 & 0.00 & 8.34  \\ 
        w/ ETO  & 35.78 & 3.47 & 0.00 & 13.08 \\ 
    \midrule
    \multicolumn{5}{c}{\textit{Llama-3.1-8B w/ Direct-Revision Trajectories}} \\
    \midrule
        \# Iter1 & 40.26 & 19.09 & 11.35 & 23.57   \\ 
        \# Iter2 & 40.35 & 27.45 & 22.97 & 30.26   \\
        \# Iter3  & 42.02 & 35.27 & 29.73 & 35.67   \\
    \midrule
    \multicolumn{5}{c}{\textit{Llama-3.1-8B w/ Trajectories from \method}} \\
    \midrule
        \# Iter1 & 46.51 & 40.82 & 32.43 & 39.92   \\  
        \# Iter2 & \underline{47.51} & \underline{45.22} & \underline{35.14} & \underline{42.62}   \\  
        \# Iter3 & \textbf{48.22} & \textbf{48.79} & \textbf{43.24} & \textbf{46.75}   \\
    \bottomrule
    \end{tabular}
    \caption{The revision results of different methods. \textit{w/ Optimal+} means the golden trajectories for training Llama-3.1-8B are generated from GPT-4o.
  }
  \label{tab:revision}
\end{table}

%% file: section/related.tex
\paragraph{Agent Learning in Interactive Environments}
Previous approaches to agent learning in interactive environments can be categorized into three strategies:
1) \textit{Prompt-based Strategy}, which uses human-written prompts to guide LLMs in summarizing experiences (\eg, constructing and refining sets of transferable skills~\citep{nottingham2024sso,sarch2024vlm} or helpful hints~\citep{chen2024automanual,majumder2023clin,zhao2024expel,yuan2024evoagent}) during exploration.
These summarized experiences are incorporated into the model's memory to enhance its knowledge and improve performance.
2) \textit{Inference-time Search Strategy}, which employs various search algorithms, such as Tree-of-Thought~\citep{yao2023tree,light2024strategist,koh2024tree,hao-etal-2023-reasoning} and Monte Carlo Tree Search (MCTS)~\citep{zhao2023large}, to identify optimal trajectories during inference. This strategy leverages prior knowledge in LLMs to enable more efficient and effective search processes~\citep{zhao2023large}.
3) \textit{Training-based Strategy}, which employs Supervised Fine-Tuning (SFT)~\citep{xi2024agentgym,qiao2024agent} or Direct Preference Optimization (DPO)~\citep{zhai2024enhancing,putta2024agent,song-etal-2024-trial,xiong-etal-2024-watch} to train LLMs. The training data can be collected from expert models~\citep{xi2024agentgym,song-etal-2024-agentbank} or generated through exploration using MCTS~\citep{zhai2024enhancing,putta2024agent,patel2024large}.

\paragraph{Self-Correction for Large Language Models}
Self-correction is a highly desirable capability of large language models (LLMs), but current LLMs have consistently been found to perform it ineffectively~\citep{xi2023rise,kamoi-etal-2024-llms,pan-etal-2024-automatically}.
Some studies on self-correcting LLMs depend on prompt-engineering~\citep{madaan2023selfrefine,shinn2023reflexion}, but often fail to achieve significant intrinsic self-correction, and can even degrade performance~\citep{huang2023large,tyen-etal-2024-llms,xie2024revealing,zhang2024understanding}.
To address these issues, several approaches propose collecting revision data from human annotators~\citep{scheurer2023training}, expert models~\citep{du2024think}, or self-generated samples~\citep{welleck2023generating,kumar2024training}. 
However, these methods primarily focus on tasks such as code repair~\citep{kim2024language,chen2024teaching}, tool use~\citep{olausson2023self}, and math~\citep{kumar2024training,havrilla2024glore}. 
These tasks provide explicit error signals, \eg, code error messages~\citep{chen2024teaching} or tool-use parameters~\citep{gou2024critic}, or rely on guidance from critical models~\citep{wang-etal-2024-math}. 
Additionally, existing methods are limited to single-turn scenarios and do not account for longer trajectories.
Although \citet{wang-etal-2024-e2cl} proposes using revision trajectories to improve agents' self-reflection in embodied tasks, it relies on expert models for correction during training data construction.
In this paper, we explore self-correction within interactive environments where language agents lack explicit error signals and rely on observations. 

%% file: section/conclusion.tex
In this paper, we address the critical challenge of error correction for language agents operating in interactive environments, a limitation that remains unresolved by existing approaches relying on expert trajectories. 
To tackle this issue, we introduce \method, an iterative self-training framework that empowers agents to dynamically reflect and correct their actions in interactive and agentic environments.
By leveraging Monte Carlo Tree Search (MCTS) and model-guided reflection construction, \method allows for timely revision of erroneous trajectories, significantly improving agents' ability to recover from errors in real time.
Experiments across three interactive environments show that \method not only enhances error correction capabilities but also prevents looping and suboptimal behaviors, leading to superior performance compared to baseline methods.
Furthermore, integrating self-generated revision trajectories into training improves the agent's overall performance and supports the development of more intelligent, self-reflective agents. 
These findings open exciting avenues for future work, particularly in refining the role of self-correction as a critical function in agent-based systems.

%% file: section/appendix.tex
\section{Prompt Template of Revision Trajectory}\label{appendix:prompt}
Table~\ref{tab:point_prompt} shows the prompt of \method to determine the transition point.

Ten different revision thoughts are shown in Table~\ref{tab:ten_thoughts}.

\section{Trajectory Definition}\label{appendix:definition}
We define four types of trajectories: initial trajectories, bad trajectories, good trajectories, and revision trajectories. 
Below, we provide formal definitions and mathematical formulations for each type of trajectory:
\begin{itemize}[leftmargin=*] 
    \item \textbf{Initial Trajectory ($\tau^{i}$)}:
    An initial trajectory represents the initial sequence of actions and observations given the user-provided instruction $u$. Formally, it is expressed as:
    \begin{equation}
    \label{eq:initial-trajectory}
    \tau^{i} = (a_{1}^{i}, o_{1}^{i}, \dots, a_{t}^{i}, o_{t}^{i}),
    \end{equation}
    where $a_{k}^{i}$ and $o_{k}^{i}$ denote the $k$-th action and the corresponding observation in the trajectory.
    
    \item \textbf{Bad Trajectory ($\tau^{b}$)}:  
    A bad trajectory extends the initial trajectory $\tau^{i}$ with a sequence of suboptimal actions and observations. It is represented as:
    \begin{equation}
    \label{eq:bad-trajectory}
    \tau^{b} = (\tau^{i}, a_{t+1}^{b}, o_{t+1}^{b}, \dots, a_{T_{b}}^{b}, o_{T_{b}}^{b}),
    \end{equation}
    where $T_{b}$ denotes the terminal step of the bad trajectory.
    
    \item \textbf{Good Trajectory ($\tau^{g}$)}:  
    A good trajectory builds upon the initial trajectory $\tau^{i}$ but involves optimal or high-reward actions and observations. It is given by:
    \begin{equation}
    \label{eq:good-trajectory}
    \tau^{g} = (\tau^{i}, a_{t+1}^{g}, o_{t+1}^{g}, \dots, a_{T_{g}}^{g}, o_{T_{g}}^{g}),
    \end{equation}
    where $T_{g}$ denotes the terminal step of the good trajectory.
    
    \item \textbf{Revision Trajectory ($\tau^{r}$)}:  
    A revision trajectory synthesizes components from both a bad trajectory $\tau^{b}$ and a good trajectory $\tau^{g}$. Specifically, it starts with the bad trajectory up to step $t'$ and transitions to the good trajectory thereafter. Formally, it is defined as:
    \begin{equation}
    \begin{aligned}
    \label{eq:revision-trajectory}
    \tau^{r} = (\tau^{i}, a_{t+1}^{b}, o_{t+1}^{b}, \dots, a_{t'}^{b}, o_{t'}^{b},\texttt{rs}, a_{t+1}^{g}, o_{t+1}^{g}, \dots, a_{T_{g}}^{g}, o_{T_{g}}^{g}),
    \end{aligned}
    \end{equation}
    where $t' \leq T_{b}$ denotes the transition point between the bad and good trajectories. $\texttt{rs}$ is the revision signal, which does not involve any environment-altering action but serves as a marker for the revision process. In our settings, \texttt{rs} is a single-turn conversation: \texttt{Assistant: [revision thought] \textbackslash n Human: OK."}. 
\end{itemize}

\input{list/point_prompt}

\input{list/ten_thoughts}

\section{Experiment Details}
\subsection{Training and Evaluation Details}\label{appendix:parameters}
All experiments are conducted using eight A100-80GB GPUs. Services for different environments are deployed on separate ports of the same server. 

The training process employed an iterative SFT approach. For the first iteration, the number of epochs is set to 3 to ensure sufficient learning. 
In subsequent iterations, the number of epochs is reduced to 1 to mitigate overfitting. 
The learning rate for iterative SFT is $2e^{-5}$, with 3\% warm-up and a cosine scheduler. 
The AdamW optimizer is used with a weight decay of 0, and a maximum gradient clipping norm of 1 is applied to prevent gradient explosion. 
The batch size is set to 1 per device, with a sequence length of 8,196 and an accumulative count of 16, adjusted based on sequence parallelism. 

Following the settings in AgentGym~\citep{xi2024agentgym}, the evaluation temperature is set to 0. 
For models not fine-tuned on expert and revision trajectories, a few-shot evaluation approach is used. 
For models fine-tuned on expert and revision trajectories, a zero-shot evaluation approach is employed.

\subsection{Multi-task training v.s. Single-task Training}\label{appendix:mix}
\input{table/MixOrNot}
We compare our multi-task training approach to single-task training, where good and revised trajectories are collected separately for each task using \method. 
The results in Table~\ref{tab:MixOrNot} demonstrate that multi-task training outperforms single-task training when using trajectories generated by \method. 
Multi-task training leverages cross-task learning to enhance performance across datasets, especially in later iterations, highlighting its scalability and effectiveness with additional training cycles. 
Trajectories from \method appear particularly beneficial for multi-task training, potentially providing richer and more diverse data for generalization. 
However, single-task training may remain advantageous for simpler or well-separated tasks, especially in earlier iterations.

\subsection{Case Study}\label{appendix:case_study}
\paragraph{Error Correction in Trajectory Generation}
As mentioned in $\mathsection$~\ref{sec:exevalu}, training with trajectories from \method significantly enhances the performance of language agents in interactive environments.
Table~\ref{table:error_correction} presents examples of historical records before and after the adaptive transition point. 
These examples demonstrate that a language agent trained on our trajectories can effectively recover from incorrect trajectories and identify the correct action.

\paragraph{Adaptive Transition Point Evaluation}
As shown in Figure~\ref{fig:revision}, we find that iterative SFT allows language agents to more easily recognize when they are on a bad trajectory with erroneous actions.
We further analyze the results of adaptive transition point evaluation to examine errors in trajectory actions. 
Examples in Table~\ref{table:point_evaluation} categorize these erroneous actions into three types:
\begin{itemize}
    \item \textbf{Invalid Actions}: Actions that are not logically permissible or executable in the current environment or context, often violating the task or system constraints.
    \item \textbf{Mismatch Observation}: Actions that contradict the observed evidence or fail to align with the current state of the environment or task requirements.
    \item \textbf{Irrelevant Actions}: Actions that do not contribute to achieving the task goal are often unrelated or misaligned with the task's specified objectives.
\end{itemize}
The examples demonstrate that language agents trained with trajectories from \method are better equipped to identify and avoid erroneous actions. 
These findings indicate that \method enhances agents' ability to differentiate between permissible, aligned, and task-relevant actions, resulting in improved decision-making and trajectory optimization in complex environments.
\input{table/error_correction}
\input{table/identify_error}

%% file: list/point_prompt.tex
\begin{table*}[t]
\small
\caption{Prompt templates of determining transition point in \method.}
    \label{tab:point_prompt}
    \centering
    \begin{tabular}{p{0.96\linewidth}}
\toprule
You are a good verifier of interactive environments. You will be given a history log that memorizes an agent interacting with the environment to solve a task. The format of the log is \\
\#\#\#\\
Action: Action \\
Observation: Observation \\
\#\#\#\\ \\

Log:
Task Description: \{\texttt{task description}\}  \\
\{\texttt{history log}\} \\
Current Action: \{\texttt{node action}\} \\
Current Observation: \{\texttt{node observation}\} \\\\

You need to verify whether the current action is good or bad or uncertain. \\
- A good action is one that is greatly helpful to solve the task. \\
- A bad action is one that is greatly harmful to solve the task.\\
- An uncertain action is one that is neither good nor bad. You cannot judge based on the current information.\\ \\
You must give reasons first and then give the response with the format: Judgment: <Good or Bad or Uncertain> \\
\bottomrule
    \end{tabular}
\end{table*}

%% file: list/ten_thoughts.tex
\begin{table*}[t]
\small
\caption{Ten revision thoughts to construct revision trajectories.}
    \label{tab:ten_thoughts}
    \centering
    \begin{tabular}{p{0.96\linewidth}}
\toprule
\textbf{Revision Thoughts} \\
\midrule
Thought 1: I realize my approach was flawed. I need to revise it. \\
Thought 2: I took the wrong actions. I need to identify the right path. \\
Thought 3: My actions were incorrect. I must adjust my strategy. \\
Thought 4: I see an error in my actions. I need to fix it. \\
Thought 5: My judgment was incorrect. I need to rethink it. \\
Thought 6: I overlooked something important. I need to address it. \\
Thought 7: I recognize my mistake. Let's find a better solution. \\
Thought 8: I recognize my failure. I need to learn and move forward. \\
Thought 9: My decision was wrong. I should reevaluate. \\
Thought 10: I made an error. I must determine how to correct it. \\
\bottomrule
    \end{tabular}
\end{table*}

%% file: table/MixOrNot.tex
\begin{table*}[t]
    \centering
    \small
    \caption{Detailed Results of different training methods on three interactive environments.}
    \label{tab:training_results}
    \begin{tabular}{lcccccc}
        \toprule
        \textbf{Dataset} & \textbf{Iteration} & \textbf{Method} & \textbf{WebShop} & \textbf{SciWorld} & \textbf{TextCraft} & \textbf{Average} \\
        \midrule
        \multirow{6}{*}{Direct-Revision + Good Trajectory} & \multirow{2}{*}{1} & Single & 48.06 & 64.52 & 57.00 & 56.53 \\
        & & Multi & 42.66 & 63.10 & 48.00 & 51.25 \\
        \cdashlinelr{2-7}
        & \multirow{2}{*}{2} & Single & 55.56 & 64.52 & 63.00 & 61.03 \\
        & & Multi & 46.93 & 64.23 & 64.00 & 58.39 \\
        \cdashlinelr{2-7}
        & \multirow{2}{*}{3} & Single & \textbf{58.82} & 65.51 & \textbf{71.00} & \textbf{65.11} \\
        & & Multi & 54.44 & \textbf{66.65} & 66.00 & 62.36 \\
        \midrule
        \multirow{6}{*}{Trajectory from \method} & \multirow{2}{*}{1} & Single & 49.80 & 64.02 & 60.00 & 57.94 \\
        & & Multi & 45.84 & 66.71 & 64.00 & 58.85 \\
        \cdashlinelr{2-7}
        & \multirow{2}{*}{2} & Single & 56.34 & 65.62 & 71.00 & 64.32 \\
        & & Multi & 52.05 & 68.28 & 76.00 & 65.44 \\
        \cdashlinelr{2-7}
        & \multirow{2}{*}{3} & Single & 60.66 & 67.22 & 76.00 & 67.96 \\
        & & Multi & \textbf{63.91} & \textbf{70.23} & \textbf{78.00} & \textbf{70.71} \\
        \bottomrule
    \end{tabular}
    \label{tab:MixOrNot}
\end{table*}

%% file: table/error_correction.tex
\begin{table*}[t]
\small
\caption{Example of error correction from the trajectory generated by Llama-3.1-8B, trained on trajectories from \method during the third iteration. The search results in WebShop are omitted due to their length.}
    \label{table:error_correction}
    \centering
    \begin{tabular}{p{0.96\linewidth}}
\toprule
\textbf{WebShop} \\
\textit{Example 1:} \\
\texttt{Assistant}: Thought: I think I should search for light blue shorts in a specific size and price range to get more accurate results. Action: search[light blue 36w 30l price:40]\\
\texttt{Human}: [Search Result]\\
\texttt{Assistant}: Thought: I went in the wrong direction with my search. I need to reset and try again.\\
\texttt{Human}: ok.\\
\texttt{Assistant}: Thought: I need to back to search. Action: click[Back to Search]\\
\texttt{Human}: [Search Result] \\ \\

\textit{Example 2:} \\
\texttt{Assistant}: Thought: I think the second option is a bit off-topic, but it's a good price, so I'll consider it. Action: click[B08226NDZW]\\
\texttt{Human}: [Search Result]\\
\texttt{Assistant}: Thought: I recognize my mistake. Let’s find a better solution. Action: wait\\
\texttt{Human}: ok.\\
\texttt{Assistant}: Thought: I think the search query is not specific enough, so I'll try to refine it by adding more keywords. Action: search[machine wash men's dress shirts polyester heathered cotton classic fit men 3t price:50.00]\\
\texttt{Human}: [Search Result] \\

\midrule
\textbf{Sciworld} \\
\textit{Example 1:} \\
\texttt{Assistant}: Thought: I should open door to kitchen. Action: open door to kitchen\\
\texttt{Human}: The door is now open.\\
\texttt{Assistant}: Thought: I took the wrong steps. I need to identify the right path. Action: wait\\
\texttt{Human}: ok.\\
\texttt{Assistant}: Thought: I should go to kitchen. Action: go to kitchen\\
\texttt{Human}: You move to the kitchen.\\ \\

\textit{Example 2:} \\
\texttt{Assistant}: Thought: I think I should try to determine the temperature of the unknown substance B. Action: use thermometer on unknown substance B\\
\texttt{Human}: the thermometer measures a temperature of 19 degrees celsius\\
\texttt{Assistant}: Thought: I misunderstood the situation. Time to reassess. Action: wait\\
\texttt{Human}: ok.\\
\texttt{Assistant}: Thought: I think I should check the unknown substance B again to make sure the temperature is accurate. Action: use thermometer on unknown substance B\\
\texttt{Human}: the thermometer measures a temperature of 16 degrees celsius \\
\midrule
\textbf{TextCraft} \\
\textit{Example 1:} \\
\texttt{Assistant}: Thought: I should check if I can get blackstone directly from the environment or the inventory. Action: inventory\\
\texttt{Human}: Inventory: You are not carrying anything.\\
\texttt{Assistant}: Thought: I recognize my mistake. Let’s find a better solution. Action: wait\\
\texttt{Human}: ok.\\
\texttt{Assistant}: Thought: I should go get the blackstone from the environment. Action: get 3 blackstone\\
\texttt{Human}: Got 3 blackstone \\ \\


\bottomrule
    \end{tabular}
\end{table*}

%% file: table/identify_error.tex
\begin{table*}[t]
\small
\caption{Example results from the evaluation of adaptive transition points. These judgments are all generated by Llama-3.1-8B, trained on trajectories from \method during the third iteration.}
    \label{table:point_evaluation}
    \centering
    \begin{tabular}{p{0.96\linewidth}}
\toprule
\textbf{WebShop} \\
\textit{Judgment (Invalid Actions):} The action is bad \dashuline{because it does not match the original search criteria}. Therefore, Judgment: Bad\\ \\

\textit{Judgment (Mismatch Observation):} However, \dashuline{the product does not match the description} of a machine wash men's dress shirt with polyester heathers, heathers cotton, cotton heather, needle sleeve, classic fit, and color heather grey, as specified in the task description. Therefore, Judgment: Bad\\ \\

\textit{Judgment (Irrelevant Actions):} The action is bad \dashuline{because it is not relevant to the task} of finding men's shirts with specific characteristics. Therefore, Judgment: Bad\\ \\
\midrule
\textbf{Sciworld} \\
\textit{Judgment (Invalid Actions):} There is no orange mentioned in the environment, so \dashuline{this action is invalid}. It's not a helpful action, but rather an invalid input. Therefore, Judgment: Bad\\ \\

\textit{Judgment (Mismatch Observation):}  The current task is to determine if the aluminum foil is electrically conductive. The agent has already connected the aluminum foil to the battery and the green light bulb, and observed that the light bulb turned on. This suggests that the aluminum foil is electrically conductive. Moving the aluminum foil to the yellow box would mean that the agent is giving up on the possibility of the aluminum foil being conductive, \dashuline{which is not in line with the current observation}. Therefore, Judgment: Bad\\ \\

\textit{Judgment (Irrelevant Actions):} The agent is not using the art studio's resources effectively, as \dashuline{the cup containing red paint is not related to the task} of creating green paint. Therefore, Judgment: Bad\\ \\
\midrule
\textbf{TextCraft} \\

\textit{Judgment (Invalid Actions):} The agent is stuck in an infinite loop of trying to execute an \dashuline{invalid action}, which is not helpful in solving the task. Therefore, Judgment: Bad\\ \\

\textit{Judgment (Mismatch Observation):} The action is not good because it is not possible to craft a dried kelp block with 9 dried kelp. \dashuline{The observation suggests that the agent lacks the necessary items to craft the block.} This action is actually harmful to the task of crafting the block. Therefore, Judgment: Bad\\ \\

\textit{Judgment (Irrelevant Actions):} The agent is trying to craft diamond boots, which requires 4 diamonds. The agent has already obtained 4 diamonds, which is sufficient to craft the boots. However, the agent is trying to craft a diamond block instead, which requires 9 diamonds. \dashuline{This action is not helpful in achieving the goal of crafting diamond boots}, as it will waste the 4 diamonds the agent already has. Therefore, Judgment: Bad\\ \\

\bottomrule
    \end{tabular}
\end{table*}